\pdfoutput=1

\documentclass[11pt]{article}

\usepackage{ACL2023} 
\usepackage{times}
\usepackage{latexsym}
\usepackage{adjustbox}

\usepackage[T1]{fontenc}

\usepackage[utf8]{inputenc}

\usepackage{microtype}

\usepackage{inconsolata}
\usepackage{amsmath}
\usepackage{booktabs}
\usepackage{graphicx}
\title{\textit{I only read it for the plot!} Maturity Ratings Affect Fanfiction Style and Community Engagement}

\author{Mia Jacobsen \\
  Center for Humanities Computing, \\ Aarhus University, Denmark \\
  \texttt{miaj@cas.au.dk} \\\And
  Ross Deans Kristensen-McLachlan \\
  Department of Linguistics,\\ Cognitive Science, and Semiotics, \\ Aarhus University, Denmark\\
  \texttt{rdkm@cc.au.dk} \\}

\begin{document}

\maketitle

\begin{abstract}
We consider the textual profiles of different fanfiction maturity ratings, how they vary across fan groups, and how this relates to reader engagement metrics. Previous studies have shown that fanfiction writing is motivated by a combination of admiration for and frustration with the fan object. These findings emerge when looking at fanfiction as a whole, as well as when it is divided into subgroups, also called \textit{fandoms}. However, maturity ratings are used to indicate the intended audience of the fanfiction, as well as whether the story includes mature themes and explicit scenes. Since these ratings can be used to filter readers and writers, they can also be seen as a proxy for different reader/writer motivations and desires. We find that explicit fanfiction in particular has a distinct textual profile when compared to other maturity ratings. These findings thus nuance our understanding of reader/writer motivations in fanfiction communities, and also highlights the influence of the community norms and fan behavior more generally on these cultural products. 

\end{abstract}

\section{Introduction}
Fanfiction is typically defined as transformational works of text that build upon an existing storyworld \cite{thomas2011fanfiction}. \textit{Fanfic}, as it is commonly known, exists in a dynamic, reciprocal relationship with the community who produce it. In one sense, fans' desires, norms, and values are shared in the form of written (generally narrative) discourse; this discourse in turn shapes the norms and values of the community over time \cite{busse2017framing, tosenberger2014mature, black2006language, evans2017more}. As such, the study of fanfiction is simultaneously the study of fans. 

A unique feature of fanfiction as a linguistic artifact is that it is regularly accompanied by community-produced metadata related to the content of the text, including proposed \textit{maturity ratings} which indicate suggested readership. 

In this study, we are interested in how the maturity ratings added by the author and used by users to filter their searches might express different reader/writer desires and motivations through their textual makeup. 

We know that fanfiction from different fandoms differ with respect to their linguistic features but are texts with more explicit maturity ratings also written differently from those suitable for general audiences? If so, what are these differences and does this constitute an \textit{explicit style}? Are there some aspects of fanfiction culture that transcend the norms of the specific communities and can be said to generalize across separate fandoms?

\subsection{Related Works}
Traditionally, research on fanfiction and fans more generally has been developed from a qualitative and ethnographic perspective \cite{barnes2015fanfiction}. These early studies showed that fanfiction writing is motivated by an \textit{admiration} for and \textit{frustration} with the source material \cite{jenkins1992textual, pugh2005democratic}. 

However, the prevalence of fanfiction texts online has led to an increasing interest in quantitative studies of fanfiction \cite{yin2017no}. The studies are often focused on either predicting the textual traits of popular or successful stories \cite{mattei2020style, nguyen2024big, sourati2022quantitative, jacobsen2024patterns}, or identifying and analyzing gender dynamics in the texts \cite{milli2016beyond, neugarten2024mythfic, yang2024exploring}. Ultimately, though, there remains a relative scarcity of literature looking to understand fanfiction as a textual phenomenon. 

Recent research from a computational perspective has provided additional evidence that writers are motivated by a complex combination of admiration and frustration \cite{jacobsen2024admiration}. Fanfic writers attempt both to imitate the source material from which they are drawn, while simultaneously preferring writing styles that break this mold in specific ways. The result is that community-preferred fanfics are less informationally dense and more focused on conversation and here-and-now interaction. In other words, fanfiction has some general genre traits upon which community-specific preferences and writing styles are super-imposed. Nevertheless, it is unclear how or how much this argument is potentially complicated by the existence of maturity ratings.

\subsection{Multidimensional Analysis}

The explicit link between the form of the text and the intention of the authors is only possible by extracting linguistic features which have concrete and readily apparent interpretations. To this end, we draw on Biber's Multidimensional Analysis (MDA) \cite{biber1988variation} to study variation across four distinct dimensions of functional variation in the English language.

With MDA, a representative excerpt of a text is tagged for presence of specific clusters of lexico-grammatical linguistic features. These features are argued to be \textit{functionally motivated}, meaning the use and prevalence of each of these features serves some kind of communicative, cognitive, or social function in the text \cite{dik1997, dik997_2, halliday2013introduction}. The distribution of these functional clusters across texts in a corpus allows us to describe the structure of texts along several \textit{dimensions of variations}.

MDA has a long history and has been widely adopted across multiple different textual registers and genres \cite{biber-1993-using-register, doi:10.1177/0075424216628955, Grieve_Woodfield_2023, jbp:/content/journals/10.1075/rs.19009.sta}; and across multiple different languages \cite{biber1995dimensions, doi:10.3366/cor.2006.1.1.1, doi:10.3366/cor.2014.0059, https://doi.org/10.1111/j.1467-971X.2009.01606.x, f514656e4d614c0c86b0efdf7a87e2ef}. Recently, the theoretical basis of MDA has been revised to include not only grammatical features but also to account for the distribution of semantically related lexical clusters, in the form of so-called \textit{Lexicalized MDA} \cite{Sardinha_Fitzsimmons-Doolan_2025}. Despite the underlying natural language processing (NLP) being somewhat basic from a contemporary perspective, MDA continues to be a robust and productive paradigm for studying variation within and across registers, not least of which is fanfiction.

In our work, we draw on the standard dimensions of variation in English regularly described by MDA \cite{biber1988variation, biber1989typology}. Table \ref{tab:mda_summary} provides a summary of some of the respective features which define these dimensions and the purpose they serve within the texts. As we will see later, the accuracy and interpretation of these labels can be questioned.

\section{Methods}
\subsection{The Corpus} 

Our corpus comprises fanfiction from three large, established fandoms based on fantasy novel series. These are Harry Potter (HP) by \citet{rowlingHP}, Percy Jackson and the Olympians (PJ) by \citet{riordanPJ}, and Lord of the Rings (LOTR) by \citet{tolkienLOTR}. The fanfics were collected from online fanfiction repository Archiveofourown.org (AO3), in accordance with their terms of service\footnote{\url{https://archiveofourown.org/tos}}. This corpus was first presented in \citet{jacobsen2024admiration}, which features a more in-depth description of the data collection process.

The corpus includes metadata from AO3, including the associated maturity ratings given by authors of the fanfic. On AO3 it is a mandatory to add a maturity rating when uploading a text to the platform. The default rating is "Not Rated" and then authors can choose to change the rating to either "General Audiences" (GA), "Teen and up Audiences" (Teen), "Mature", and "Explicit". According to AO3's FAQ, the ratings are based on the following definitions\footnote{\url{https://archiveofourown.org/faq/tags}}: 

\begin{quote}
\textbf{General Audiences}   The content is unlikely to be disturbing to anyone, and is suitable for all ages. \\
\textbf{Teen And Up Audiences}  The content may be inappropriate for audiences under 13.\\
\textbf{Mature} The content contains adult themes (sex, violence, etc.) that aren't as graphic as explicit-rated content.\\
\textbf{Explicit} The content contains explicit adult themes, such as porn, graphic violence, etc. 
\end{quote}

We excluded any fanfic tagged with \textbf{Not Rated} as we wanted texts where the author and reader both made intentional choices as to the content of the text. The final corpus is summarized in Table \ref{tab:summary_stats}.

\begin{table}[!ht]  
    \centering
    \caption{Summary of the corpus}
    \begin{tabular}{l|cccc}
     & HP & PJ & LOTR \\
    \midrule
    \textbf{GA} & 51,441 & 6,888 & 6,315 \\
    \textbf{Teen} & 74,779 & 3,261 & 7,128\\
    \textbf{Mature} & 45,606 & 1,465 & 1,636 \\
    \textbf{Explicit} & 48,799 & 1,838 & 1,488\\
     \midrule
    Total & 220,625 & 12,879 & 17,140 \\
    \end{tabular}
    \label{tab:summary_stats}
\end{table}

\begin{figure*}[!htb]
    \centering
    \includegraphics[width=1\textwidth]{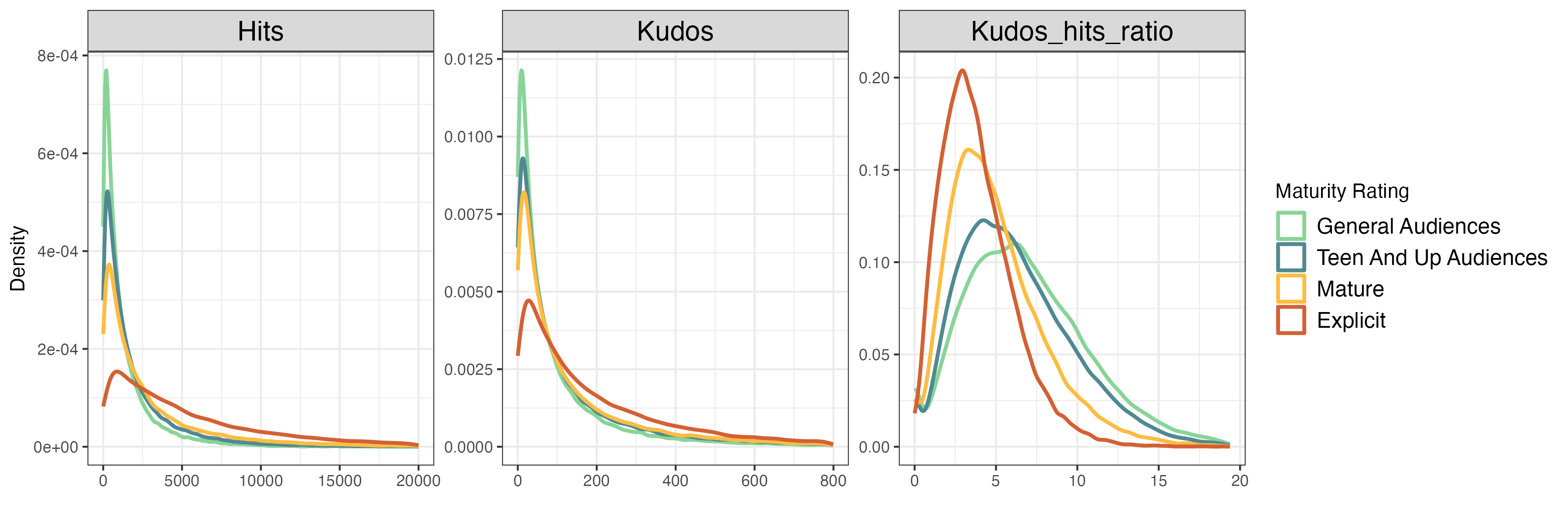}
    \caption{Density distributions of three engagement metrics across maturity ratings. Hits is the number of times a fanfiction has been opened, kudos is analogous to the number of likes, and the kudos/hits ratio is the number of kudos divided by the number of hits times 100.}
    \label{fig:engagment_distributions}
\end{figure*} 

\begin{table*}[!t]
    \centering
    \caption{Summary of dimensions of variation established using MDA. Modified from 
    \citet{jacobsen2024admiration} and
     \citet{nini2019multi}}
    \begin{adjustbox}{width=1\textwidth}

    \begin{tabular}{l | p{0.2\linewidth} | p{0.4\linewidth} | p{0.3\linewidth}}
        & \textbf{Summary} & \textbf{Short Description } & \textbf{Examples of Features}\\
        \midrule
        D1 & Involved / Informational Discourse & \textit{Informational}: Dense and careful information integration. \newline \textit{Involved}: Affective and intertactional style, like conversations & \textit{Informational}: type/token-ratio, prepositions, nouns \newline \textit{Involved}: first and second person pronouns, contractions, present tense, emphatics\\
        & & \\
        D2 & Narrative Concern & Distinguishes between texts with a narrative focus from others & Past tenses, third person pronouns, perfect aspects, public verbs\\
        & & \\
        D3 & Context-(in)dependent Referents & \textit{Context-dependent}: Receiver must use context to infer what time and place is being referred to. \newline \textit{Context-independent}: The referents in the text are made explicit and thus not dependent on the context & 
        \textit{Context-dependent}: time adverbials, place
        adverbials, general adverbs \newline \textit{Context-independent}: wh- relative clauses on object position, wh- relative clauses on subject position, nominalizations\\
        & & \\
        D4 & Overt Expression of Persuasion & The degree to which the sender’s opinion is overtly expressed and/or overt attempts to persuade the receiver are made & Infinitives, prediction
        modals, suasive verbs, necessity modals\\
    \end{tabular}
    \end{adjustbox}
\label{tab:mda_summary}
\end{table*}

Using the same feature extraction and statistical method as \citet{jacobsen2024admiration}, we wish to characterize the textual profiles of fanfiction texts with different maturity ratings to further understand the fans and their motivation for reading and writing fanfiction. In that particular paper, reader engagement metrics are modeled directly on the dimension scores based on Biber's MDA, with no consideration given to the effect of maturity ratings on the relationship between dimension scores and reader engagement.

However, when looking at the engagement metrics for the different maturity ratings, a clear and perhaps somewhat surprising pattern emerges. On Figure \ref{fig:engagment_distributions}, density distributions for three different engagement metrics often employed in studies of fanfiction are visualized for each of the maturity ratings. The three engagement metrics are, respectively, the number of hits (i.e., the number of times a fanfic has been opened by a user), the number of kudos (i.e., the number of likes), and the kudos/hits ratio used in (i.e., the number of kudos divided by the number of hits - referred to subsequently as the \textit{K-H ratio}). 

The figure shows that across maturity ratings, \textbf{Explicit} fanfiction generally has a \textit{lower} K-H ratio compared to the other ratings. This is despite \textbf{Explicit} fanfiction being a popular and appreciated genre as visible on the distributions for hits and kudos, where the rating lies above the others as the numbers increases, especially for hits. 

The K-H ratio is intended to balance the raw number of hits and kudos for a fanfiction, with the goal of removing or minimizing the effect of time and general popularity. However, as shown on Figure \ref{fig:engagment_distributions}, it can be seen how it devalues the appreciation for \textbf{Explicit} fanfiction. Since a fanfic can only receive one kudos per user but multiple hits upon revisits, \textbf{Explicit} fanfics generally have a lower K-H ratio simply because they are revisited more. This is problematic inasmuch as it introduces bias into most studies on the style of popular and successful fanfiction texts, especially as \textbf{Explicit} fanfiction texts constitute a substantial amount of fanfiction of the corpus, as illustrated in Table \ref{tab:summary_stats}. 

This dynamic in the engagement metric motivated the current study to add nuance to the way quantitative studies conceptualize the writing style of popular or successful fanfiction, as the role of the fans and their desires need to be accounted for. As such, this study focuses on understanding how the norms of fan communities influence how fanfiction is written.

\subsection{Feature Extraction}
As Biber's original MDA method is not publicly released, we used the Multidimensional Analysis Tagger (MAT) as developed by Andrea Nini \cite{nini2019multi}. Nini's MAT is based on the grammatical features as described in \citet{biber1988variation}. 

The tagger takes a corpus of text excerpts and tags them for each of the included linguistic features. Afterwards, it uses the prevalence of the different features to score each text on each of the dimensions of functional variation. 

This means that for each fanfiction, we have a score for the degree of \textit{Involved versus informational discourse} (D1), the degree of \textit{Narrative Concern }(D2) in the text, the degree of \textit{Context-(in)dependent Referents} (D3), and the degree of \textit{Overt Expression of Persuasion} (D4). Dimensions 5 and 6 were excluded, as their robustness and usefulness for fanfiction has been questioned \cite{jacobsen2024admiration}.

Although this is a dictionary-based approach, we argue that the value in functionally motivated features and the subsequent clear understanding of \textit{why} the fanfiction texts might be written in this way up-weighs the downsides one might otherwise see with dictionary-based approaches.

\subsection{Statistical Analysis}
For the statistical analysis, we created a series of linear mixed effects models to test for the effect of maturity ratings and fandom on the different dimension scores. Linear mixed effects models are a useful tool in this specific case, as these types of models perform in robust and predictable ways even with imbalanced data \cite{snijders2011multilevel, METEYARD2020104092}.

Additionally, since one author can be in the dataset multiple times if they have posted multiple fanfics that fit the search criteria, a regular linear regression is not possible, as it will violate the assumption of independence of data points. Mixed effects models instead offer a way to explicitly model the fact that authors can occur multiple times in the dataset by adding random intercepts. As such, they account for these repeated measures when estimating the effects.

Using the package lmerTest \cite{lmertest2017} for R \cite{Rlang2023}, we created a linear mixed effects model for each of the four dimensions of variation, which sought to predict the dimension scores for the given dimension from an interaction between the fandom (HP/LOTR/PJ) and the maturity rating (GA/Teen/Mature/Explicit). 

\begin{figure*}[!htb]
    \centering
    \includegraphics[width=1\textwidth]{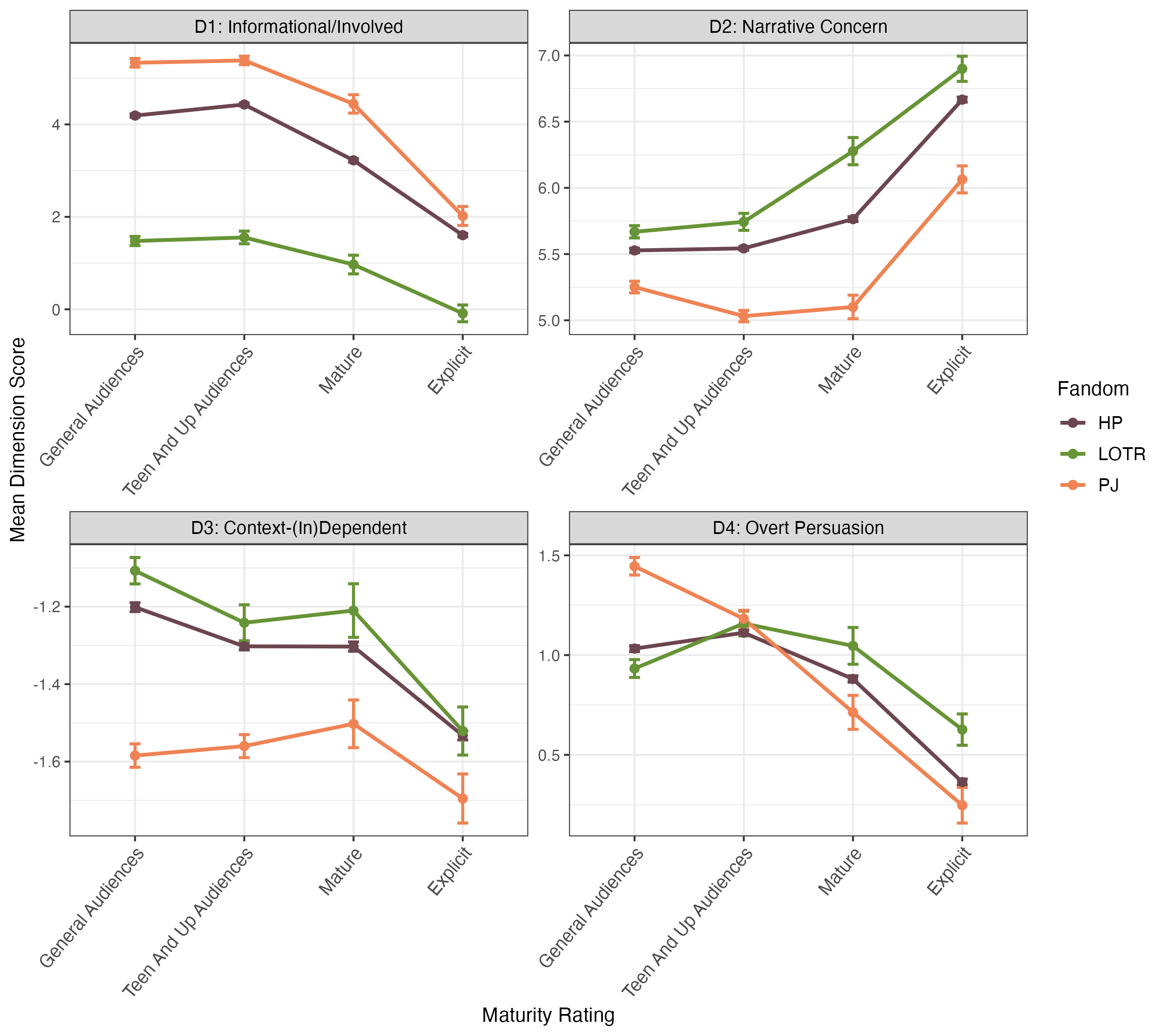}
    \caption{Mean and standard error for the dimension scores across maturity ratings and fandoms.}
    \label{fig:dim_scores_ratings}
\end{figure*} 

Word counts and publication dates were scaled and added to the models as control variables. A random intercept was added for author. The model therefore looked as follows:

\begin{equation}
\begin{split}
  Dimension \sim maturity\ rating\ *\ fandom\ +\\ word\ count \ +\ published\ date\ +\ (1 | author)
\end{split}  
\end{equation}

This means that for each of the four dimension of variation, the estimated difference between maturity ratings across fandom groups will be found. 

\section{Results}
The findings are visualized on Figure \ref{fig:dim_scores_ratings}, which shows the mean dimension score for each maturity rating across fandoms. A regression table showing the specific outputs from the models can be seen in Table \ref{tab:model_outputs_d1}.

From a visual inspection, there is a clear change from \textbf{GA} to \textbf{Explicit} for each of the dimensions, which is present across all three fandoms. Most strikingly, the \textbf{Explicit} group looks quite distinct in its textual profile compared to the other groups. The following model summaries allow us to disentangle these visual patterns more definitively. 

\begin{table*}[!htp]
\centering
\caption{Estimates for model (1) for each dimension of variation}
    \begin{tabular}{ l | c c c c }
     \textbf{Dimension 1}& $\beta$& SE &  t-value & p-value \\ 
      \midrule
     Teen & 0.18 & 0.046 & 3.84 & < 0.001*\\
     Mature & -0.84 & 0.054 & -15.56 & < 0.001*\\
     Explicit & -2.39 & 0.057 & -42.86 & < 0.001*\\
     LOTR & -2.47 & 0.13 & -18.5 & < 0.001* \\  
     PJ & 1.35 & 0.12 & 11.57 &  < 0.001* \\
     Teen:LOTR & -0.35 & 0.18 & -1.94 & 0.052 \\
     Mature:LOTR & -0.018 & 0.24 & 0.072 & 0.94 \\
     Explicit:LOTR & 0.84 & 0.24 & 3.58 & < 0.001* \\
     Teen:PJ & -0.21 & 0.14 & -1.44 & 0.15 \\
     Mature:PJ & 0.093 & 0.23 & 0.40 & 0.69 \\
     Explicit:PJ & -0.31 & 0.25 & -1.22 & 0.22 \\
    \midrule
     \textbf{Dimension 2} & $\beta$& SE &  t-value & p-value \\ 
      \midrule
     Teen & 0.023 & 0.022 & 1.05 & 0.29\\
     Mature & 0.34 & 0.025 & 13.44 & < 0.001*\\
     Explicit & 1.15 & 0.026 & 43.79 & < 0.001*\\
     LOTR & 0.24 & 0.063 & 3.84 & < 0.001* \\  
     PJ & -0.18 & 0.55 & -3.29 &  < 0.01* \\
     Teen:LOTR & 0.16 & 0.09 & 1.85 & 0.065 \\
     Mature:LOTR & 0.29 & 0.11 & 2.55 & < 0.05* \\
     Explicit:LOTR & 0.081 & 0.11 & 0.73 & 0.46 \\
     Teen:PJ & -0.13 & 0.068 & -1.91 & 0.056 \\
     Mature:PJ & -0.26 & 0.11 & -2.39 & < 0.05* \\
     Explicit:PJ & -0.15 & 0.12 & -1.29 & 0.20 \\
     \midrule
      \textbf{Dimension 3} & $\beta$& SE &  t-value & p-value \\ 
      \midrule
     Teen & -0.10 & 0.015 & -6.72 & < 0.001*\\
     Mature & -0.17 & 0.017 & -9.63 & < 0.001*\\
     Explicit & -0.37 & 0.018 & -20.51 & < 0.001*\\
     LOTR & 0.10 & 0.041 & 2.49 & < 0.05* \\  
     PJ & -0.34 & 0.037 & -9.36 &  < 0.001* \\
     Teen:LOTR & -0.011 & 0.059 & -0.19 & 0.85 \\
     Mature:LOTR & 0.034 & 0.079 & 0.43 & 0.67 \\
     Explicit:LOTR & -0.029 & 0.075 & -0.38 & 0.70 \\
     Teen:PJ & 0.083 & 0.047 & 1.78 & 0.075 \\
     Mature:PJ & 0.11 & 0.074 & 1.52 & 0.13 \\
     Explicit:PJ & 0.040 & 0.081 & 0.49 & 0.62 \\
     \midrule
      \textbf{Dimension 4}& $\beta$& SE &  t-value & p-value \\ 
      \midrule
     Teen & 0.013 & 0.020 & 0.67 & 0.51\\
     Mature & -0.19 & 0.023 & -8.44 & < 0.001*\\
     Explicit & -0.68 & 0.023 & -28.68 & < 0.001*\\
     LOTR & -0.12 & 0.054 & -2.13 & < 0.05* \\  
     PJ & 0.30 & 0.048 & 6.21 &  < 0.001* \\
     Teen:LOTR & 0.12 & 0.078 & 1.58 & 0.11 \\
     Mature:LOTR & 0.15 & 0.10 & 1.4772 & 0.14 \\
     Explicit:LOTR & 0.25 & 0.099 & 2.49 & < 0.05* \\
     Teen:PJ & -0.13 & 0.062 & -2.15 & < 0.05* \\
     Mature:PJ & -0.34 & 0.098 & -3.43 & < 0.001*\\
     Explicit:PJ & -0.16 & 0.11 & -1.50 & 0.13 \\

    \end{tabular}
\label{tab:model_outputs_d1}
\end{table*}     

The findings for \textit{Involved versus Informational Discourse} (D1), shows five significant main effects and one significant interaction effect. With \textbf{GA} fanfics as a baseline, \textbf{Teen} fanfics are slightly more involved, while \textbf{Mature} fanfics are more informational, and \textbf{Explicit} fanfics are the most informational. The one significant interaction effect shows that LOTR has a slightly smaller difference between \textbf{GA} and \textbf{Explicit} fanfics as compared to HP and PJ. Worth noting then is that the general pattern of change in maturity ratings remains similar across groups despite the fandoms having significantly different levels of \textit{Involved/Informational Discourse} (D1).

The model for the second dimension, which describes the degree of \textit{Narrative Concern} (D2) in the texts, shows four significant main effects and two significant interaction effects. For the main effects for maturity ratings, compared to \textbf{GA}, \textbf{Teen} shows no difference, whereas \textbf{Mature} has a slightly higher degree of narrative concern and \textbf{Explicit} fanfiction has the greatest degree of narrative concern. 

Looking at the interaction effects, there is again generally the same pattern of change in maturity ratings across fandoms. The only exceptions occur in the \textbf{Mature} category for PJ and LOTR, which compared to HP, respectively, have a greater degree and lesser degree of \textit{Narrative Concern} when compared to their respective \textbf{GA} fanfics.

For the third dimension, \textit{Context-(in)dependent referents}, we find only significant main effects and no interaction effects. This means that although the different fandoms have distinct levels of context-dependence, 
across the maturity ratings the degree of change is similar. As the maturity ratings go from \textbf{GA} to \textbf{Teen}, \textbf{Mature}, and \textbf{Explicit}, so do the referents in the texts become more context-dependent. This means that fanfics become more here-and-now oriented. 

This is surprising, as more context-dependent referents (low D3 score) are typically associated with a more involved style (high D1 score) \cite{nini2019multi} but we find the opposite pattern across maturity ratings.

For the fourth and final dimension, \textit{Overt Expression of Persuasion}, we find four main effects and three interaction effects. For the maturity ratings, there is no difference between \textbf{GA} and \textbf{Teen} fanfics. \textbf{Mature} fanfics, however, have less overt persuasion than \textbf{GA}, and \textbf{Explicit} continues that trend with the least overt persuasion. 

The interaction effects indicate that these patterns are slightly dependent on the fandom. Specifically, for LOTR, the \textbf{Explicit} group has a positive interaction effect meaning less difference between \textbf{GA} and \textbf{Explicit} than for HP. For PJ, there are two significant interaction effects. These show that, compared to \textbf{GA} fanfics in PJ, \textbf{Teen} and \textbf{Mature} show even less overt persuasion than \textbf{Teen} and \textbf{Mature} from HP and LOTR. 

So, in contrast to the other dimensions where the change across ratings was similar, we find that the different maturity ratings in PJ have a quite different change in \textit{Overt Expression of Persuasion} (D4) than the other two fandoms. 

\section{Discussion}
These findings indicate that although general preferences can be found across fandoms, what is desired from one's fanfiction is quite dependent on the flavor of fanfiction that is sought out by the reader. 

\textbf{Explicit} fanfiction is so clearly distinct from the other three maturity ratings in ways that, for the most part, are similar across groups. This particular result alone adds significant nuance to the established conception of fans' desires \cite{jacobsen2024admiration, nguyen2024big, sourati2022quantitative}, both as writers and readers. Specifically, the general focus on characters and their interpersonal relationships are still generally present, but the way these interactions are characterized changes drastically dependent on the genre of fanfiction. The writing style of the fanfics are thus not only dependent on the source material of the specific fandom. Instead, there are norms that transcend the individual community as to how specific "genres" are to be written, regardless of the specific fandom.

For \textbf{Explicit} fanfiction, the greater information presentation is situated within the story's context which is subsequently what creates the unique combination of dimension scores, i.e., both informational discourse and context-dependent referents. The texts are descriptive and action-focused but not necessarily meant to drive a plot or be carefully planned. The action and the descriptions are focused on the here-and-now, indicating that character interaction is still the main focus of these texts, but the way character interactions can be focal to a story is not only confined to dialogue. In these cases, the actions speak louder than the words. 

These findings also call for a nuanced interpretation of the different labels for the four dimensions of variation. \textbf{Explicit} fanfiction is not typically known to be a genre that is, for example, plot-driven, which one might otherwise expect based on the greater degree of \textit{Narrative Concern} (D2) within the texts. 

In their overview of so-called \textit{pornographic transformative works}, \citet{joseph2024pornography} not only show the myriad of ways fans re-contextualize the source material, they also highlight that \textbf{Explicit} or pornographic fanfiction often has a lesser focus on plot. This is sometimes known within fandom as PWP fics or "Porn without Plot" / "Plot, What Plot?" fanfics \cite{joseph2024pornography}, highlighting how both readers and writers of fanfiction go into texts well knowing what to expect.

As such, Biber's \textit{Narrative Concern} (D2) does not necessarily only cover "narrative" in the classic sense of plot and story structure. What this study shows is that these dimensions also lend themselves to further interpretation. For example, Dimension 2 can also be understood as a focus on character movements and actions. 

\section{Conclusion}
Together, this study paints a picture of \textbf{Explicit} fanfiction as standing out from those with lower maturity ratings. It appears to be a genre of its own with a conventional focus on descriptions, actions, and here-and-now orientation. The patterns of dimension scores found for \textbf{Explicit} fanfiction are unusual in that they combine features that are not usually correlated in earlier work. 

\textbf{Explicit} fanfiction thus nuances the findings from previous quantitative studies that take a more general look at fanfiction. While it is true that fans in general might prefer fanfiction stories with a more involved style and less narrative focus, the different maturity ratings show us that fans' motivation for reading and writing fanfiction is as much colored by the source material they build upon as it is on the distinct genre of fanfiction they wish to contribute with.

When taken together with the bias that engagement metrics might incorporate towards \textbf{Explicit} fanfiction, it is crucial that future research take these dynamics into account when making statements about the writing style of successful or popular fanfiction.

\section{Limitations}
This paper has focused on a small subsection of available fanfiction. All three fandoms included in the study center around Western media, specifically fantasy novel series. As such, the analysis could benefit from a wider and less Western gaze on fanfiction to better understand the genre as a whole. Especially since this analysis has shown fan communities have distinct preferences and norms.

Additionally, as mentioned, Biber's MDA is a dictionary-based approach, meaning that findings are generally confined to what is included in the list of features compiled by Biber and subsequently incorporated into the MAT created by Nini. This means that a great deal of contextual and general knowledge is missing. This kind of world-knowledge is something which readers of fanfiction undoubtedly make use of from a cognitive stylistic perspective when reading and engaging with the texts \cite{emmott_narrative_1997, gerrig1993experiencing, herman2004story,sanford_emmott_2012}. Taking into account the community-specific language that is typical in fan communities, more contextual features could provide further insight into the specific dynamics of fanfiction. 

Finally, although this study criticizes the bias potentially introduced by the K-H ratio and other engagement metrics, there is no statistical analysis to support this argument. It can be argued that although these maturity ratings differ in writing style, the general writing style of, say, \textbf{Explicit} fanfiction, might not be the most preferred within the communities. In other words, a prevalent style is not necessarily an appreciated one. Further research is needed to more deeply understand the interactions between fan preferences and the way it influences fanfiction writing. 

\section{Ethics Statement}
This study builds upon a corpus of publicly available texts obtained from the AO3 platform that was collected in accordance with the terms of service outlined on their website\footnote{\url{https://archiveofourown.org/tos##I.E}}. However, we recognize that for fanfiction there is an added responsibility pertaining to data stewardship. Fanfiction texts often deal with personally sensitive topics pertaining to identity markers as gender and sexuality, as well as (re)tellings of traumatic experiences which the fanfiction is written to help process. 

While many members of the platform adopt pseudonyms, it is nevertheless true that, in the case of quantitative studies of this size that build upon online data, it is not possible to obtain ethical consent from the fanfiction authors. Additionally, there is the added complexity of copyright as it pertains to the authors of the source material.

With these considerations in mind, we opted to ensure that our research data was treated as personally sensitive information. It was stored in accordance with European GDPR legislation and the access was limited to only the authors of this paper. As the analysis in this paper is limited to text-level features that are focused on the form rather than the content of the texts and removed from any specific user, any negative impact on specific users should be mitigated.

\section{Acknowledgements}
Part of the computation done for this project was performed on the UCloud interactive HPC system, which is managed by the eScience Center at the University of Southern Denmark. 
  
We also wish to thank the readers and writers of fanfiction, particularly those who contribute to AO3.

\bibliography{anthology, custom}
\bibliographystyle{acl_natbib}

\appendix

\end{document}